\documentclass{article}

\usepackage[preprint]{neurips_2021}

\usepackage[utf8]{inputenc} 
\usepackage[T1]{fontenc}    
\usepackage{hyperref}       
\usepackage{url}            
\usepackage{booktabs}       
\usepackage{amsfonts}       
\usepackage{nicefrac}       
\usepackage{microtype}      
\usepackage{xcolor}         

\usepackage[acronym]{glossaries}

\DeclareMathAlphabet{\mathcal}{OMS}{cmsy}{m}{n}

\usepackage{hyperref}
\usepackage{url}
\usepackage{amsmath}
\usepackage{amsthm}

\usepackage{amssymb}
\usepackage{multirow}
\usepackage{subcaption}
\usepackage{bm}

\usepackage{amsmath,amsfonts,bm}

\def\eqref#1{equation~\ref{#1}}

\def\1{\bm{1}}

\def\vx{{\bm{x}}}

\DeclareMathAlphabet{\mathsfit}{\encodingdefault}{\sfdefault}{m}{sl}
\SetMathAlphabet{\mathsfit}{bold}{\encodingdefault}{\sfdefault}{bx}{n}

\def\gD{{\mathcal{D}}}

\def\gX{{\mathcal{X}}}

\newcommand{\softmax}{\mathrm{softmax}}

\DeclareMathOperator*{\argmax}{arg\,max}
\DeclareMathOperator*{\argmin}{arg\,min}

\usepackage{url}

\usepackage{color}
\usepackage{graphicx}
\usepackage{float}
\usepackage{subcaption}
\usepackage{multirow}
\usepackage{siunitx}
\newcommand{\norm}[1]{\Vert#1\Vert}

\newcommand{\abs}[1]{\vert#1\vert}

\newcommand{\bq}{\begin{equation}}
\newcommand{\eq}{\end{equation}}

\newcommand{\prob}{\mathbb{P}}
\newcommand{\Prob}[2][{}]{\prob_{#1}\left[#2\right]}
\DeclareMathOperator{\TopW}{{Top-1}}

\DeclareMathOperator{\ttrain}{{train}}
\DeclareMathOperator{\ttest}{{test}}
\DeclareMathOperator{\tcal}{{cal}}
\DeclareMathOperator{\tin}{{in}}
\DeclareMathOperator{\tout}{{out}}
\newcommand{\trainset}{\gD^{\ttrain}_{\tin}}
\newcommand{\testset}{\gD^{\ttest}_{\tin}}
\newcommand{\calset}{\gD^{\tcal}_{\tin}}
\newcommand{\calseti}{\gD^{\tcal,i}_{\tin}}
\newcommand{\calsetj}{\gD^{\tcal,j}_{\tin}}
\newcommand{\oodset}{\gD^{\ttest}_{\tout}}

\newcommand{\pmax}{{p^{\max}}}

\def\1{\bm{1}}

\usepackage{xcolor}

\usepackage{bbding}
\usepackage{pifont}

\newcommand{\good}{\color{green}{\ding{52}}}
\newcommand{\bad}{\color{red}{\ding{56}}}

\title{Frustratingly Easy Uncertainty Estimation for Distribution Shift}

\author{%
  Tiago Salvador\\
  Mila, McGill University\\
  \And
  Vikram Voleti \\
  Mila, Universit\'{e} de Montr\'{e}al\\
  \AND
  Alexander Iannantuono \\
  UBC Okanagan \\
  \And
  Adam Oberman \\
  Mila, McGill University\\
}

\begin{document}

\maketitle
\newacronym{dnn}{DNN}{Deep Neural Network}
\newacronym{ood}{OoD}{Out-of-Distribution}

\begin{abstract}
Distribution shift is an important concern in deep image classification, produced either by corruption of the source images, or a complete change, with the solution involving domain adaptation. While the primary goal is to improve accuracy under distribution shift, an important secondary goal is uncertainty estimation: evaluating the probability that the prediction of a model is correct. While improving accuracy is hard, uncertainty estimation turns out to be frustratingly easy. Prior works have appended uncertainty estimation into the model and training paradigm in various ways. Instead, we show that we can estimate uncertainty by simply exposing the original model to corrupted images, and performing simple statistical calibration on the image outputs. Our frustratingly easy methods demonstrate superior performance on a wide range of distribution shifts as well as on unsupervised domain adaptation tasks, measured through extensive experimentation.
\end{abstract}

\section{Introduction}

As deep learning models become more ubiquitous, it has become increasingly critical to estimate their predictive uncertainty i.e. how reliable their predictions are.
This is particularly important in healthcare, financial, and legal settings where a human user makes a decision aided by a deep learning model.
The predictive uncertainty of a deep classification model is typically provided by an estimate of the class-wise probabilities of a given sample.

The baseline predictive uncertainty method is to simply use the softmax probabilities of the model as a surrogate for the class-wise probabilities~\citep{hendrycks17baseline} (we will refer to this as the \textbf{Vanilla} method).
However, such probability estimates are known to lead to overconfident models~\citep{nguyen_deep_2015}, and several approaches have been proposed to calibrate these probabilities. These methods include non-Bayesian ones such as \textbf{Temperature Scaling} (TS)~\citep{guo_calibration_2017}, dropout \citep{srivastava2014dropout,gal2016dropout}, and model ensembles \citep{lakshminarayanan_simple_2017}, as well as Bayesian approaches such as Stochastic Variational Bayesian Inference (SVBI) for deep learning \citep{graves2011SVI,blundell2015weightuncertainty,louizos2016SVI,louizos2017SVI,wen2018flipout}, among others. 

All these approaches produce calibrated probabilities for in-distribution data with varying degrees of success. However, it has been found that the quality of the uncertainty predictions deteriorates significantly for data under \textit{distributional shift} \citep{hendrycks2019robustness, SnoekPaper}.

\citet{SnoekPaper} demonstrate this by performing a large-scale benchmark analysis of existing methods for predictive uncertainty under dataset shift. In particular, CIFAR-10 and ImageNet trained models were evaluated on (i) synthetic shifts using their corrupted counterparts CIFAR-10-C and ImageNet-C \citep{hendrycks2019robustness}.
Other types of dataset shifts include (ii) natural shifts, such as from ImageNet to ImageNet-v2~\citep{ImageNetV2} or ImageNet-Sketch~\citep{ImageNetSketch}, and (iii) domain shift, from synthetic images to real images, such as in Office-Home~\citep{OfficeHome}, or VisDA-2017~\citep{peng2017visda}.

\begin{table}[t]
\caption{Comparison of different calibration methods.}
\centering
\setlength{\tabcolsep}{.3em}
\begin{tabular}{l|cccc}\toprule
Calibration & works with & no & no & no target \\
\:\:\:Methods & domain shift & features &  retuning & images\\\midrule
Vanilla & \bad & \good & \good & \good\\
TS & \bad & \good & \good & \good\\
CPCS & \good & \bad & \bad & \bad\\
TransCal & \good & \bad & \bad & \bad\\
\textbf{SAC} (Ours) & \good & \good & \good & \bad\\
\textbf{STS} (Ours) & \good & \good & \good & \good\\
\bottomrule
\end{tabular}
\label{table:properties_comparison}
\vspace{-1em}
\end{table}

In this work, we propose two frustratingly easy and effective post hoc methods for model calibration under distributional shift : Surrogate Adaptive Calibration (SAC), and Surrogate Temperature Scaling (STS). Our focus is on calibration for unsupervised distribution shift, where only unlabeled images of the test distribution are available. Our methods provide superior results among comparable methods, on the three types of distribution shift mentioned above: (i) synthetic shifts (CIFAR-10-C, ImageNet-C), (ii) natural shifts (ImageNet-V2, ImageNet-Sketch), and (iii) domain shifts (Office-Home, VisDA-2017).
The key idea is that we can estimate distribution shift using just the model outputs, by comparing with known (arbitrary) corruption of in-distribution data.

Our methods can be combined with training-based calibration methods, and show impressive results on never seen types of corruptions. Our methods require no intermediate features from the model, and do not need any retraining, retuning, or additional training of models.

\section{Related Work}

Models trained on a given dataset are unlikely to perform as well on a shifted data \citep{hendrycks2019robustness, SnoekPaper}, and there are inevitable tradeoffs between accuracy and robustness \citep{chun2020empirical}. Several approaches have been proposed to increase model robustness, typically evaluated on the benchmark corrupted datasets CIFAR-10-C and ImageNet-C. Methods such as
\citet{hendrycks2019pretraining} show that fine-tuning a pre-trained model can improve the quality of the uncertainty estimates compared to training a model from scratch, but it does not improve accuracy. In contrast, we propose simple calibration methods that require no additional training or fine-tuning.

Simply training models against corruptions can fail to make models robust to new corruptions \citep{vasiljevic2016examining, geirhos2018generalisation}. 
However, \citet{hendrycks2020augmix} train models with a carefully designed new data augmentation technique called AUGMIX, and are able to improve both robustness and uncertainty measures.
In contrast, our work applies a data augmentation technique at the calibration stage, which avoids having to retrain the models from scratch, and can be a simple post-hoc fix to calibrate trained models.

\citet{nado2020evaluating} argue that the internal activations of deep models also suffer from distributional shift in the presence of shifted data. They thus propose to recompute the batch normalization coefficients at prediction time using a sample of the unlabeled images from the test distribution. While the accuracy, and ultimately the calibration, was improved for synthetic shifts (ImageNet-C), it degraded the accuracy for natural shifts (ImageNet-v2). Moreover, their work requires knowing the internal activations of the model, and hence is a white-box calibration which works only on deep neural networks. In contrast, we improve any model's calibration treating it as a black box, and hence do not require to know its internal functionalities.

\citet{AvUCloss} introduced a new loss function,
and demonstrated that it can be used as post-hoc calibration method leading to improved accuracy and uncertainty measures against new corruptions. This work can be seen as orthogonal to ours as the methods we propose here can be used as an add-on to any post-hoc supervised calibration method.

\citet{shao2020calibrating} propose a confidence calibration method that uses an auxiliary classifier to identify mis-classified samples, thus allowing them to be assigned low confidence. However, it requires a small sample of labeled test data. In contrast, our method does not require target labels.

While the above works focus on synthetic shifts modeled by noise corruptions, natural shifts (e.g. clipart to real images) are of greater interest.
These are prominent in unsupervised domain adaptation, where only unlabeled examples from the test distribution are available.
Many algorithms have been proposed to leverage these during training, which can be split into two main groups: (a) \emph{moment matching}, which look to minimize the Maximum Mean Discrepancy \citep{NIPS2012MKMMD} between source (labeled) and target (unlabeled) images at the feature level (e.g. DAN \citep{Long15DAN}, JAN \citep{Long17JAN}); and (b) \emph{adversarial training}, inspired by the success of Generative Adversarial Networks (GANs) \citep{Goodfellow14GAN}: DANN \citep{Ganin2016DANN} introduced a domain discriminator that differentiates source features from target features, competing with the feature extractor;
CDAN \citep{Long18CDAN} improved upon DANN by conditioning the model on the classifier predictions;
SAFN \citep{Xu2019SAFN} observe that the accuracy degradation from source to target model arises from the smaller feature of the target domain in comparison to the source domain, and propose an Adaptive Feature Norm approach to fix it;
MDD \citep{Zhang2019MDD} proposed a new domain adaptation margin theory;
MCC \citep{Jin2020MCC} proposed a new loss to minimize class confusion i.e. the tendency of the classifier to confuse correct and ambiguous classes for the target examples. However, these methods only focus on accuracy, and do not leverage calibration for uncertainty estimation.

Domain adaptation algorithms improve accuracy at the expense of model calibration \citep{wang2020transferable}. To fix this, \cite{ParkUDA} proposed \textbf{CPCS}, an approach based on importance weighting to correct for covariate shift in data by deriving an upper bound on the expected calibration error. \textbf{TransCal}~\citep{wang2020transferable} extended the Temperature Scaling method into domain adaption, and achieved more accurate calibration with lower bias and variance, without introducing any hyperparameters.
(Concurrent work by \cite{Gupta2020} presents some preliminary simulations also using importance weighting, but limited to Random Forest and binary classifiers.)

However, both CPCS and TransCal require access to the model's features to train an additional domain classifier, which is then used to compute the importance weights. In contrast, our methods only require the softmax probabilities of the model, thereby treating the model as a black-box, and can be extended to classifiers other than neural networks.
Moreover, CPCS and TransCal need to be re-tuned for every new corruption to compute new importance weights. In contrast, our calibration methods require no retraining of any sort, and performs calibration for each surrogate set a priori.

\section{Calibration}

\subsection{Supervised Calibration}

Consider the $K$-class classification problem, where  $\vx \in \gX$ is a set of inputs such as images, and $y \in \{1, \ldots , K\}$ denotes the corresponding labels. The inputs and labels are drawn i.i.d. from the joint distribution  $p(\vx, y)$, and $y$ is a sample from the conditional distribution $p(y \mid \vx)$.

A classifier $f(\vx)$ is trained using a training dataset $\trainset$, with hyper-parameters selected using a validation/calibration dataset $\calset$. The datasets $\trainset$ and $\calset$ consist of finite samples drawn i.i.d. from $p(\vx, y)$.

Typically when $f$ is a deep neural network, $f : \gX \to [0,1]^{K}$ has a terminal softmax layer applied to linear outputs $g(\vx)$, i.e. $f(\vx) = \softmax(g(\vx))$. Hence, the model $f$ outputs a probability distribution on the $K$ labels given an input $\vx$ from $\gX$ (our analysis can be generalized to the case where the model outputs scores). The predicted class $\hat y$ is given by the most likely output:
\begin{align}
\hat y(\vx) = \underset{k}{\argmax} f(\vx)_k
; \quad
\pmax(\vx) = \max_k f(\vx)_k
\end{align}
The model confidence (or uncertainty) $c_k$ for label $k$ is defined as
the probability that the true label is $k$ given the classifier's softmax output for that label $f(\vx)_k$:
\begin{align}
\label{eq:confidence_class}
c(\vx;p,f)_k = \Prob[p(\tilde{\vx},y)]{y=k \mid f(\tilde{\vx})_k = f(\vx)_k}
\end{align}
We write $c(\vx;p,f)$ to emphasize $c$'s dependence on both $f$ and $p(\vx, y)$, since the distribution will change below. We shorten the notation whenever it is clear from context.

The goal of \textit{supervised calibration} \citep{ParkUDA} is to estimate the calibrated probabilities $c$ empirically by $\hat{c}$ using a finite set of labeled samples $\calset$ drawn from $p(\vx, y)$.  
The error between the true and the estimated confidences is typically measured by Expected Calibration Error (ECE$\downarrow$) \citep{guo_calibration_2017}:
\begin{align}
\label{eq:calib_error}
\text{ECE} = \mathbb{E}_{p(\vx, y)}\left[ \norm{c(\vx) - \hat{c}(\vx)}\right]
\end{align}

\textbf{Vanilla} : $\hat c(\vx) = f(\vx)$, i.e. the \textit{Vanilla} approach simply estimates ${c}(\vx)$ as the softmax probabilities $f(\vx)$.

In general, these softmax probabilities are not an accurate prediction of the class probabilities \citep{domingos1996beyond}. In particular, for deep neural network models they are overconfident predictions \citep{guo_calibration_2017}. They become even more overconfident under distribution shift \citep{SnoekPaper}.

\textbf{Temperature scaling}: $\hat{c}(\vx) = \text{softmax}(g(\vx)/T)$, and an optimal $T$ is estimated \citep{guo_calibration_2017} by minimizing the negative log likelihood loss on the calibration set $\calset$.

\subsection{Covariate Calibration for Unsupervised Domain Adaptation}

Unsupervised domain adaptation is the case where in addition to labeled images $\trainset$ from the source distribution $p(\vx, y)$, we also have unlabeled images $\oodset$ from a target distribution $q(\vx, y)$ that is different from $p$. Here, $q$ is said to have a covariate shift from $p$, i.e. $q(y | \vx) = p(y | \vx)$ but $q(\vx) \neq p(\vx)$ \citep{Shimodaira2000}. The standard way of training models involves using images only from $p$. However, domain adaptation methods may leverage the unlabeled images $\oodset$ from $q$ during training.

The goal of \textit{covariate calibration}~\citep{ParkUDA} is to estimate calibrated probabilities $c$ for $q$:
\begin{align}
\label{eq:confidence_class_covariate}
c(\vx;q,f)_k = \Prob[q(\tilde{\vx},y)]{y=k \mid f(\tilde{\vx})_k = f(\vx)_k}
\end{align}
The challenge lies in the fact that while we are given a calibration dataset of labeled examples $\calset$ drawn from $p$, we only have a dataset of unlabeled examples $\oodset$ drawn from $q$.
In other words, \autoref{eq:confidence_class_covariate} requires labels to estimate $c$, which are not available.

\textbf{Temperature Scaling} (TS) does not tackle this directly, and assumes that the calibrated probabilities for $p$ must work well on $q$.

\textbf{CPCS}~\citep{ParkUDA} and \textbf{TransCal}~\citep{wang2020transferable} compute an importance weight for each sample in $\calset$ using samples from $\trainset$ and $\oodset$, and then use a weighted version of temperature scaling:
\begin{enumerate}
    \item Compute $w(\vx) = q(\vx)/p(\vx)\ \forall x \in \calset$, using $\oodset$
    \item Optimize $T$ in temperature scaling by minimizing a weighted calibration error.
\end{enumerate}
\textbf{CPCS}~\citep{ParkUDA} minimizes the Brier Score~\citep{degroot1983comparison} weighted by $w$.  \textbf{TransCal}~\citep{wang2020transferable} minimizes ECE weighted by $w$, with additional terms to correct for bias and variance in the expectations.

CPCS and TransCal need to compute importance weights to estimate the distribution shift, and then calibrate through temperature scaling. This unfortunately implies that new weights and temperatures need to be computed for every new type of corruption, and every new intensity.

\begin{figure*}[t]
\centering
\includegraphics[width=\linewidth]{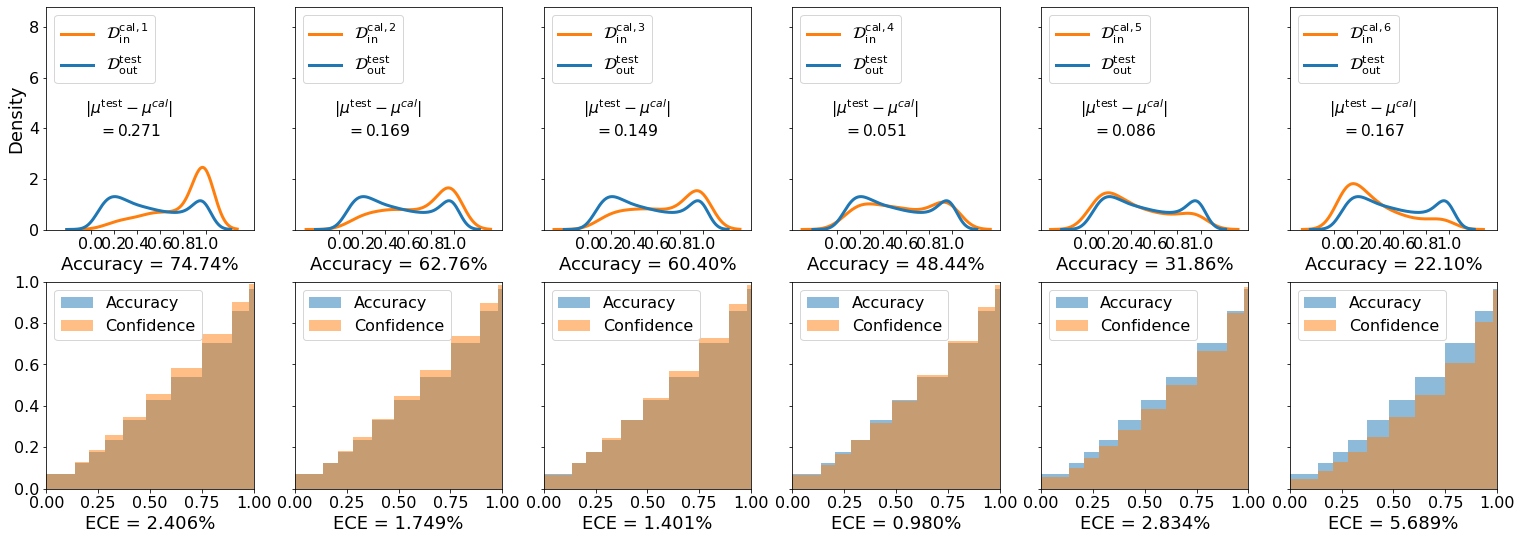}
\vspace{-.5em}
\caption{TOP: Probability density (using kernel density estimation) of $\oodset$ (blue) obtained by corrupting ImageNet test images with the ``elastic transform'' at intensity 2, and the $\pmax$ distribution of each calibration set $\calsetj$ (orange) obtained by corrupting ImageNet-cal images with varying intensity of a different corruption ``pixelate''.
BOTTOM: The respective accuracy and calibration confidence. The minimum calibration error is achieved precisely when the means of the distributions are the closest.
}
\label{fig:imagenet_surrogate_distributions}
\end{figure*}

\section{Our methods}

We conjecture that \emph{distribution shift can be imitated by model output shift} for calibration. While the former is a complex distribution shift, the latter is one dimensional and can be handled by simple statistical techniques.
Our solution is to find a surrogate shifted dataset (where the labels are known), with a $\pmax$ model distribution similar to the unknown one. Then, we can use this surrogate dataset to calibrate the unknown one in a supervised fashion by simply setting $c(\vx;q,f)_k = c(\vx;q^*,f)$, where $q^*$ denotes the surrogate dataset distribution.

We empirically observe that these values indeed match, provided the $\pmax(\tilde \vx)$ distribution for $\tilde \vx \sim q^*$ is close to the $\pmax(\vx)$ distribution for $\vx \sim q$. \autoref{fig:imagenet_surrogate_distributions} TOP shows the $\pmax$ distributions of corrupted ImageNet-test set, and six surrogate calibration sets synthesized from the ImageNet-calibration set by adding increasing levels of a different corruption. It can be seen that the 4th surrogate distribution represents the closest match with the test distribution. Indeed, in \autoref{fig:imagenet_surrogate_distributions} BOTTOM, we see that this corresponds to the least calibration error.

Our approach is motivated by the theory on domain adaptation by \citet{TheoryDA2007,TheoryDA2010} which suggests that successful transfer across distributions is one where the model cannot identify to which distribution the input observation belongs to. We achieve precisely this by matching with a surrogate distribution.

\textbf{Choice of supervised calibration method}:
We first choose a supervised calibration method, and adapt it for covariate calibration. We chose Temperature Scaling \citep{guo_calibration_2017}, and show that this simple calibration method is sufficient to estimate uncertainty reliably. Note that any other calibration method could be used instead of Temperature Scaling as part of our methods.

While more sophisticated supervised calibration methods have recently been proposed \citep{Kull2019beyond,Rahimi2020intraorder}, Temperature Scaling remains a competitive baseline.
Moreover, CPCS \citep{ParkUDA} and TransCal \citep{wang2020transferable}, which are the most relevant prior methods for covariate calibration, both utilize Temperature Scaling. This makes it ideal for a fair comparison.

\subsection{Surrogate Adaptive Calibration (SAC)}

\textbf{Synthesize surrogate calibration sets}: Given a set of finite samples $\calset$ drawn from $p(\vx, y)$ not seen during the training of the model, we form $J$ distinct calibration sets $\calsetj$, $j=\{1,\ldots,J\}$, by corrupting the data with a known corruption at different levels of intensity.
This step is equivalent to drawing samples from distributions $q^j(\vx, y)$, $j=\{1,\ldots,J\}$ with a covariate shift from $p(\vx, y)$.
However, unlike the unlabelled data drawn from $q$, the labels of $q^j$ are known. Therefore, we can apply supervised calibration methods on each $q^j$ to obtain confidence estimates.
The final uncertainty estimate is then
provided by the best $q^j$ according to some metric. We expand on the number of surrogate calibration sets and choice of corruption in Section \ref{sec:variants}.

\textbf{Calibrate the surrogate sets}: For each calibration set $\calsetj$, we define by $T_j$ the optimal temperature obtained through Temperature Scaling. Thus, recaling that $g$ denotes the model's scores pre-softmax, we have $\hat{c}(\vx;q^j,f) = \softmax(g(\vx)/T_j)$.

Then, given a test set of samples $\oodset = \{\vx_1, \dots, \vx_m\}$ drawn from $q(\vx, y)$,
\begin{enumerate}
	\item Calculate the mean of the $\pmax$ values of the test images:
\begin{align}
    \mu^{\ttest} = \mu(\pmax(\oodset)) = \frac{1}{\abs{\oodset}}\sum_{\vx \in \oodset}\pmax(\vx)
\end{align}
	\item Record the surrogate calibration set with the mean of its $\pmax$ values i.e. $\mu^{\tcal,j} = \mu(\pmax(\calsetj))$
closest to $\mu^{\ttest}$:
    \begin{align}\label{eq:zs}
        i = \underset{j}{\argmin} |\mu^{\ttest} - \mu^{\tcal,j}|
    \end{align}
	\item Calibrate $\oodset$ according to the surrogate calibration set with the closest mean:
    \begin{align}
        \hat{c}(\vx;q,f) = \hat{c}(\vx;q^i,f)
    \end{align}
\end{enumerate}

Our calibration method is simple, and computationally efficient. The most expensive step is in evaluating the calibration images and respective corruptions with the classifier. This can be done efficiently since the number of images is relatively low. By relying on Temperature Scaling, which requires solving a one parameter optimization, the supervised calibration on each surrogate set can also be computed quickly. These can be computed and stored a priori. Therefore, at the evaluation stage we only need to compare the means.

\subsection{Surrogate Temperature Scaling (STS)}

We also propose a (frustratingly) easy and effective variant of SAC. Instead of choosing one surrogate calibration set $\calseti$ with the closest mean $\pmax$ to the test set, we calibrate according to the union of all the synthesized surrogate calibration sets $\calsetj$. We find that this performs better than the prior methods, and just as effectively as SAC.

We compare our methods' characteristics with those of relevant prior methods in Table \ref{table:properties_comparison}. 

\section{Evaluation Metric}

As is tyically the case, we evaluate calibration error in terms of the probability of \emph{correct classification} i.e. Top-1 correctness:
\begin{equation}
\label{eq:confidence_Top1_covariate}
c^{\TopW}(\vx;q,f) = \Prob[q(\tilde{\vx}, y)]{ y = \hat y(\tilde{\vx}) \mid \pmax(\tilde{\vx}) = \pmax(\vx)}
\end{equation}
Our estimated Top-1 confidence estimate is naturally given by:
\begin{align}
\hat{c}^{\TopW}(\vx;q,f) = \max_k \hat{c}^{\TopW}(\vx;q,f)_k.
\end{align}
We then measure Top-1 ECE, by binning the data based on the probability of the most probable class according to the classifier i.e. $\pmax(\vx)$.
For each bin $B_m$, $m\in\{1,\ldots,M\}$, we estimate the true Top-1 model confidence by: 
\begin{align}
\label{eq:binned_accuracy}
c^{\TopW}(B_m) = \frac{1}{\abs{B_m}}\sum_{(\vx,y)\in B_m}\1_{\hat{y}(\vx)=y}
\end{align}
where $\1_{\hat{y}(\vx) = y}$ is 1 if $\hat{y}(\vx) = y$, else 0. 

Similarly, the empirical bin model confidence is: 
\begin{align}
\label{eq:binned_confidence}
\hat{c}^{\TopW}(B_m) = \frac{1}{\abs{B_m}}\sum_{(\vx,y)\in B_m}\hat{c}^{\TopW}(\vx;q,f)
\end{align}
Then, the bin ECE is calculated as the binned version of Eq. \ref{eq:calib_error}, 
the weighted-average of the absolute difference between ${c}^{\TopW}(B_m)$ and $\hat{c}^{\TopW}(B_m)$:
\begin{align}
\label{eq:ece}
ECE = \sum_{m=1}^M \frac{\abs{B_m}}{N}\abs{c^{\TopW}(B_m)-\hat{c}^{\TopW}(B_m)}
\end{align}
where $N$ is the total number of test samples, and $\abs{B_m}$ is the number of samples in bin $B_m$. The actual value of ECE depends on the binning procedure used: equally spaced, or equally sized.
Equal spacing leads to bins with very few samples since deep learning models have high accuracy and therefore, if calibrated, high confidence. To mitigate this issue, we use equal sized bins similar to \citet{adaptiveECE} and \citet{hendrycks2019oe}.
In our experiments, we use $M=15$ equal-sized bins.

\section{Datasets}

\subsection{Synthetic shifts}

We evaluate on synthetic distribution shift using the CIFAR-10 and ImageNet datasets, and their corrupted counterparts CIFAR-10-C and ImageNet-C \citep{hendrycks2019robustness}.
The latter were formed by applying common real-world corruptions (16 in total) at 5 levels of intensity to the 10,000 test images of CIFAR-10, and the 50,000 test images of ImageNet.
Corruptions include brightness (variations in daylight intensity), Gaussian noise (in low-lighting conditions) and Defocus blur (when the image is out of focus).
See the Appendix for examples of each of the corruptions used.
We chose 5,000 images to form our $\calset$. The remaining images form $\testset$, and their corrupted versions form $\oodset$.

\subsection{Natural shifts}

We evaluate on natural distribution shift using:
\begin{enumerate}
    \item ImageNet-V2~\citep{ImageNetV2} dataset, which was designed to be as similar as possible to the original ImageNet dataset, but the data collection process took place a decade after ImageNet. It contains three different versions resulting from different sampling strategies: Matched-Frequency (MF), Threshold-0.7 (Thr), Top-Images (TI).
    \item ImageNet-Sketch~\citep{ImageNetSketch} dataset, which shares the same 1000 classes as ImageNet but all the images are black and white sketches.
\end{enumerate}

\subsection{Domain Adaptation}

In the context of Domain Adaptation, we consider:
\begin{enumerate}
    \item Office-Home~\citep{OfficeHome}, a medium-scale dataset with 65 classes and 4 domains: Artistic (A), Clipart (C), Product (P) and Real-World (R). Models are trained using labeled images of one domain (source) and unlabeled image of a different domain (target). We consider all 12 combinations of domain shifts.
    \item VISDA-2017, a large-scale dataset with a 12-class object recognition task and a large domain shift from synthesis-to-real images.
    The training set contains 152,000 synthetic images generated by rendering 3D models, while the test set has 55,000 real object images sampled from Microsoft COCO~\citep{MicrosoftCOCO}.
\end{enumerate}

We consider two different partitions of each domain depending on if it is used as a source or target distribution. 
When used as the source distribution, we do a 80/20 split into $\trainset$ and $\calset$. When used as a target distribution, we do a 50/50 split. One split is used during training, and the other forms our $\oodset$. We present calibration on $\oodset$. 

\section{Experimental details}
\label{sec:exp}

\subsection{Relevant comparisons}

Our proposed method performs post-training confidence calibration for unsupervised domain adaptation. Thus we compare our SAC and STS methods to CPCS~\citep{ParkUDA} and TransCal~\citep{wang2020transferable}. As additional baselines, we also consider Vanilla (softmax probabilities) and Temperature Scaling (TS)~\citep{guo_calibration_2017}.

\subsection{Models}

\textbf{Source-only}:
For the CIFAR10 models, we used the ResNet-20 pretrained model provided in the Github repo ``Sandbox for training deep learning networks''\footnote{https://github.com/osmr/imgclsmob}~\citep{imgclsmob}. During training, the images were randomly distorted using horizontal flips and random crops and their brightness, contrast and saturation were jittered.

For the ImageNet models, we used pretrained ResNet-50 provided by PyTorch, which used the same data augmentation during training. Hence, brightness, contrast and saturation, together with the corruption used in our SAC and STS methods, are removed from CIFAR10-C and ImageNet-C when testing.

\textbf{Domain Adaptation}:
We trained models using the Transfer Learning Library Github repo\footnote{https://github.com/thuml/Transfer-Learning-Library}~\citep{dalib}. The different domain adaptation methods we used include DAN~\citep{Long15DAN}, JAN~\citep{Long17JAN}, CDAN~\citep{Long18CDAN}, MDD~\citep{Zhang2019MDD}, SAFN~\citep{Xu2019SAFN}, MCC~\citep{Jin2020MCC}.

\subsection{Choices in SAC and STS}
\label{sec:variants}

Our calibration methods SAC and STS require the synthesis of multiple calibration sets of varying corruption intensity from the original calibration set. We used ``pixelate'' as the corruption, and generated $J=6$ calibration sets.

\textbf{Choice of surrogate sets}:
We form $J=6$ calibration sets $\calsetj$, since CIFAR10-C and ImageNet-C contain 5 levels of corruption intensity. We chose to include the original calibration set, and 5 surrogate sets with increasing corruption intensity. For $j=1$, we simply take the images in $\calset$. For $j>1$, we take their ``pixelate''-corrupted counterparts with intensity level $j-1$.

\textbf{Choice of corruption for calibration}:
We find that the choice of corruption for calibration does not affect the overall performance significantly. We performed a cross-validation study over the choice of corruption used to generate the calibration sets (always leaving it out of the corruptions used at test time). \autoref{fig:ECEcorruption_analysis} (in the appendix) plots the mean and variance of the ECE across different choices for calibration corruptions.

It is to be noted that the corrupted images at test time have \textit{never been seen by the model}, either at the training stage or the calibration stage. Despite this, our easy calibration methods perform quite well. The $\oodset$ of the synthetically shifted CIFAR-10-C and Imagenet-C are formed by perturbing the images in the respective $\testset$ with corruptions different from those used during training and calibration, i.e. not brightness, contrast, saturation, or pixelate. Hence, $\calsetj$ and $\oodset$ are completely disjoint. Despite this, our calibration shows improved results on $\oodset$ in the cases of covariate shift.

\textbf{Alternative distances}: In Eq. \ref{eq:zs}, instead of the distance between the means, any other difference between distributions could be used. We tried the Kolmogorov-Smirnoff statistic, and the Wasserstein distance between the cumulative distributions of the $\pmax$ values  (see \autoref{fig:multi_image_distance} in the appendix), and do not find significant change in performance from using the mean.

\textbf{More efficient mean calculation}: Instead of computing the mean of the full $\oodset$, we computed it for a random subset of only 100 samples, and found similar performance (see \autoref{fig:multi_image_peak} in the appendix). As both CPCS and TransCal require substantially higher samples for an effective estimation of the importance weights, this is a significant advantage of our SAC method.

\section{Results}
\label{sec:results}

\begin{figure}
\centering
\includegraphics[width=0.5\linewidth]{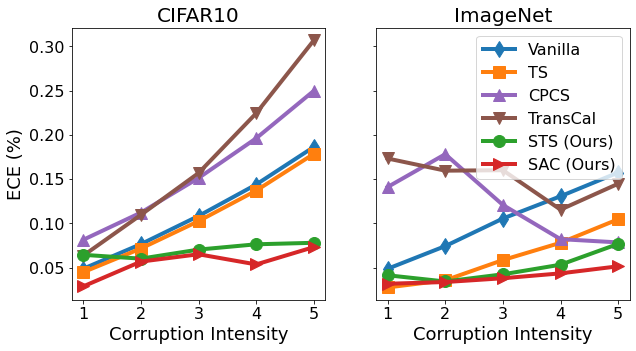}
\caption{Mean ECE \% (lower is better) across corruption types (synthetic distribution shift) for 5 corruption intensities using different calibration methods: Vanilla, Temperature Scaling (TS), CPCS \citep{ParkUDA}, TransCal \citep{wang2020transferable}, our Surrogate Temperature Scaling (STS), and our Surrogate-Adaptive Calibration (SAC).
Our methods perform the best among all methods across all intensities, with the greatest improvement at higher intensities.}
\label{fig:ECE_means}
\end{figure}

\subsection{Synthetic shift}
We applied the different calibration methods mentioned before to the CIFAR10 and ImageNet trained models, and evaluated their calibration on CIFAR10-C and ImageNet-C. We report the ECE means across the different corruptions in \autoref{fig:ECE_means}.
For a more detailed comparison, we have included in the appendix (see \autoref{fig:ECE_boxwhisker}) a box plot with different quantiles summarizing the ECE results for each method and intensity shift.
Importantly, we leave out brightness, contrast and saturation corruption since they were used during training as well as the pixelate corruption used by both STS and SAC method during calibration.

Both figures show that SAC and STS \textit{consistently outperform} all other methods. They minimize ECE the most across all levels of corruption intensity, with even better performance at higher levels of intensity.

CPCS and TransCal are able to improve calibration at higher corruption intensities only for the ImageNet model. Since the models are trained using only the source distribution, there is a large discrepancy between source (clean images) and target (corrupted images) distributions, which leads to bad importance weight estimation, and ultimately poor calibration by CPCS and TransCal.

\subsection{Natural shift}
The results of evaluating the calibration methods on all three versions of ImageNet-V2 and ImageNet-Sketch are displayed in \autoref{table:imagenet_natural_shift}. Once again, SAC and STS provide the best results.

\begin{table}
\centering
\caption{ECE \% (lower is better) for natural distribution shifts on ImageNet.}
\label{table:imagenet_natural_shift}
\begin{tabular}{l|ccc|c}
    & \multicolumn{3}{c|}{ImageNet-V2} & \multirow{2}{*}{ImageNet-Sketch} \\
    &  MF &  Thr &  TI &  \\
\midrule
Vanilla     & 8.77 & 4.88 & 3.23 & 22.29 \\
TS          & 4.81 & 2.49 & 1.87 & 16.09 \\
CPCS        & 5.29 & 3.26 & 5.85 & 38.79 \\
TransCal    & 12.26 & 4.43 & 1.99 & 52.24 \\
\textbf{SAC} (Ours)  & \textbf{4.44} &  2.49 & \textbf{1.86} & \textbf{10.71} \\
\textbf{STS} (Ours)  & 4.71 & \textbf{2.39} & 1.90 & 15.86 \\
\end{tabular}
\end{table}

\subsection{Domain Adaptation}

For the Office-Home dataset, we trained a different model for each of the 12 combinations of domain shifts, using the CDAN adaptation algorithm \citep{Long18CDAN}. We applied the different calibration methods on the split of the target distribution not seen during training, so as to mimic a more challenging case and close to real-world application. The results are displayed in \autoref{table:cdan_officehome_unseen}. Our simple STS method provides the best calibration in 6 out of the 12 domain shifts.

As for the VisDA-2017 dataset, which contains only one domain shift from synthetic to real images, we trained different models using 6 different domain adaptation methods: DAN \citep{Long15DAN}, JAN \citep{Long17JAN}, CDAN \citep{Long18CDAN}, MDD \citep{Zhang2019MDD}, SAFN \citep{Xu2019SAFN}, MCC \citep{Jin2020MCC}. The results are shown in \autoref{table:visda_unseen}. We find that our
STS method provides the best calibration error, followed by our SAC method.

\subsection{Ablation study on SAC}
Our SAC method is based on two key components: using corrupted data for calibration, and adaptive calibration based on the model's mean confidence. In order to measure the contribution of each component we can compare SAC to the STS and TS. The results show that STS improves upon TS at higher levels of corruption intensity, thus proving the effectiveness of surrogate calibration sets. However, this comes at the cost of poorer calibration at low corruption intensities. The SAC method not only provides bigger improvements in calibration at higher corruption intensities, it also retains calibration at lower intensities. This highlights the importance of the adaptive mechanism.

\textbf{Why it works}:
Our results are based on the fact that we can use the model outputs, in our case simply the $\pmax$ values, as a proxy to evaluate distribution shift.
This was illustrated in \autoref{fig:imagenet_surrogate_distributions}: the calibration set whose $\pmax$ mean is closest to the $\pmax$ mean of the test set typically has lower calibration error.
This is precisely what our SAC method does.

\begin{table*}
\centering
\caption{ECE \% (lower is better) comparison on target data not seen during training on \textit{Office-Home} with different domain adaptation methods.}
\label{table:cdan_officehome_unseen}
\resizebox{\textwidth}{!}{
\begin{tabular}{l|ccccccccccccc}
{} &  A2C &  A2P &  A2R &  C2A &  C2P &  C2R &  P2A &  P2C &  P2R &  R2A &  R2C &  R2P &   Avg \\
\midrule
Vanilla    &  27.88 &  17.42 &   6.70 &  13.26 &  13.07 &  10.81 &  12.75 &  26.07 &   5.45 &   7.49 &  20.90 &   5.62 & 13.95 \\
TS         &  28.08 &  17.88 &   8.39 &  11.09 &  10.97 &   8.72 &  15.07 &  27.06 &   6.10 &   6.32 &  21.37 &   5.63 & 13.89 \\
CPCS       &  18.18 &  17.33 &   9.36 &  10.22 & \textbf{10.21} &  15.88 &  10.18 &  34.26 &  11.38 &   7.42 &  17.12 &   5.12 & 13.89 \\
TransCal   & \textbf{10.76} &  16.26 &  14.65 &  12.50 &  11.32 & \textbf{6.38} & \textbf{9.56} & \textbf{22.34} &   5.80 &   7.09 & \textbf{9.51} &   8.85 & \textbf{11.25}\\
\textbf{SAC} (Ours) &  28.08 &  17.88 &   8.39 &  11.14 &  10.89 &   8.14 &  17.72 &  29.44 &   7.41 &   6.56 &  20.54 &   5.63 & 14.32 \\
\textbf{STS} (Ours) &  25.23 & \textbf{15.26} & \textbf{4.86} & \textbf{10.2} &  11.47 &   7.90 &  14.55 &  27.97 & \textbf{5.28} & \textbf{6.03} &  15.95 & \textbf{4.94} & 12.47 \\
\bottomrule
\end{tabular}}
\end{table*}

\begin{table}
\centering
\caption{ECE \% (lower is better) on target data not seen during training on \textit{VisDA-2017} with different domain adaptation methods.}
\label{table:visda_unseen}
\setlength{\tabcolsep}{.25em}
\begin{tabular}{lcccccc}
\toprule
{} &   DAN &   JAN &  CDAN &   MDD &   SAFN &   MCC \\
\midrule
Vanilla    & 21.10 & 27.46 & 18.44 & 23.34 & 22.93 & 23.58 \\
TS         & 19.24 & 29.85 & 21.03 & 24.56 & 19.55 & 24.03 \\
CPCS       &  6.67 & 31.04 & 21.89 & 24.17 & 25.78 & 24.53 \\
TransCal   & 25.61 & 34.48 & 22.81 & 25.88 & 29.73 & 24.84 \\
\textbf{SAC} (Ours) &  7.83 & 23.66 & 18.68 & 22.72 & \textbf{7.05} & 23.32 \\
\textbf{STS} (Ours) & \textbf{5.33} & \textbf{20.03} & \textbf{9.04} & \textbf{18.11} &  9.45 & \textbf{22.38}\\
\bottomrule
\end{tabular}
\end{table}

\section{Conclusions}

Increasingly, models trained on a given dataset are being asked to perform on data from a ``shifted'' dataset.
Our work focuses on calibration of uncertainty estimates: we wish to ensure the model's output probability reflects the true probability of the event. In contrast to most deep uncertainty work, we use a (frustratingly easy) purely statistical approach, and successfully reduce the calibration error of deep image classifiers under dataset shift.

We propose 2 methods, where we add a simple extra calibration step, and estimate uncertainty using only the model outputs. Our calibration methods involve synthesizing surrogate calibration sets by simply adding increasing intensities of a single type of known corruption to the original calibration data. while one method leverages unlabeled target images to chose the best surrogate set, the other simply uses their union.
Our methods only require the output probabilities of a model, unlike previous methods which typically require the features of a model. Moreover, our methods do not require any retraining or retuning or additional training.

While previous works have shown that uncertainty estimates degrade on corrupted data,
our two proposed calibration methods lead to better calibration on various types of dataset shifts: synthetic, natural, and domain adaptation. This is evidenced quantitatively in extensive experiments.
\bibliographystyle{plainnat}
\bibliography{neurips_2021}

\begin{thebibliography}{49}
\providecommand{\natexlab}[1]{#1}
\providecommand{\url}[1]{\texttt{#1}}
\expandafter\ifx\csname urlstyle\endcsname\relax
  \providecommand{\doi}[1]{doi: #1}\else
  \providecommand{\doi}{doi: \begingroup \urlstyle{rm}\Url}\fi

\bibitem[Ben-David et~al.(2007)Ben-David, Blitzer, Crammer, and
  Pereira]{TheoryDA2007}
Shai Ben-David, John Blitzer, Koby Crammer, and Fernando Pereira.
\newblock Analysis of representations for domain adaptation.
\newblock In B.~Sch\"{o}lkopf, J.~Platt, and T.~Hoffman, editors,
  \emph{Advances in Neural Information Processing Systems}, volume~19. MIT
  Press, 2007.
\newblock URL
  \url{https://proceedings.neurips.cc/paper/2006/file/b1b0432ceafb0ce714426e9114852ac7-Paper.pdf}.

\bibitem[Ben-David et~al.(2010)Ben-David, Blitzer, Crammer, Kulesza, Pereira,
  and Vaughan]{TheoryDA2010}
Shai Ben-David, John Blitzer, Koby Crammer, Alex Kulesza, Fernando Pereira, and
  Jennifer~Wortman Vaughan.
\newblock A theory of learning from different domains.
\newblock \emph{Machine Learning}, 79\penalty0 (1):\penalty0 151--175, 2010.
\newblock \doi{10.1007/s10994-009-5152-4}.
\newblock URL \url{https://doi.org/10.1007/s10994-009-5152-4}.

\bibitem[Blundell et~al.(2015)Blundell, Cornebise, Kavukcuoglu, and
  Wierstra]{blundell2015weightuncertainty}
Charles Blundell, Julien Cornebise, Koray Kavukcuoglu, and Daan Wierstra.
\newblock Weight uncertainty in neural network.
\newblock In Francis Bach and David Blei, editors, \emph{Proceedings of the
  32nd International Conference on Machine Learning}, volume~37 of
  \emph{Proceedings of Machine Learning Research}, pages 1613--1622, Lille,
  France, 07--09 Jul 2015. PMLR.
\newblock URL \url{http://proceedings.mlr.press/v37/blundell15.html}.

\bibitem[Chun et~al.(2020)Chun, Oh, Yun, Han, Choe, and Yoo]{chun2020empirical}
Sanghyuk Chun, Seong~Joon Oh, Sangdoo Yun, Dongyoon Han, Junsuk Choe, and
  Youngjoon Yoo.
\newblock An empirical evaluation on robustness and uncertainty of
  regularization methods.
\newblock \emph{arXiv preprint arXiv:2003.03879}, 2020.

\bibitem[DeGroot and Fienberg(1983)]{degroot1983comparison}
Morris~H DeGroot and Stephen~E Fienberg.
\newblock The comparison and evaluation of forecasters.
\newblock \emph{Journal of the Royal Statistical Society: Series D (The
  Statistician)}, 32\penalty0 (1-2):\penalty0 12--22, 1983.

\bibitem[Domingos and Pazzani(1996)]{domingos1996beyond}
Pedro Domingos and Michael Pazzani.
\newblock Beyond independence: Conditions for the optimality of the simple
  bayesian classi er.
\newblock In \emph{Proc. 13th Intl. Conf. Machine Learning}, pages 105--112,
  1996.

\bibitem[Gal and Ghahramani(2016)]{gal2016dropout}
Yarin Gal and Zoubin Ghahramani.
\newblock Dropout as a bayesian approximation: Representing model uncertainty
  in deep learning.
\newblock In Maria~Florina Balcan and Kilian~Q. Weinberger, editors,
  \emph{Proceedings of The 33rd International Conference on Machine Learning},
  volume~48 of \emph{Proceedings of Machine Learning Research}, pages
  1050--1059, New York, New York, USA, 20--22 Jun 2016. PMLR.
\newblock URL \url{http://proceedings.mlr.press/v48/gal16.html}.

\bibitem[Ganin et~al.(2016)Ganin, Ustinova, Ajakan, Germain, Larochelle,
  Laviolette, March, and Lempitsky]{Ganin2016DANN}
Yaroslav Ganin, Evgeniya Ustinova, Hana Ajakan, Pascal Germain, Hugo
  Larochelle, Fran{\c{c}}ois Laviolette, Mario March, and Victor Lempitsky.
\newblock Domain-adversarial training of neural networks.
\newblock \emph{Journal of Machine Learning Research}, 17\penalty0
  (59):\penalty0 1--35, 2016.
\newblock URL \url{http://jmlr.org/papers/v17/15-239.html}.

\bibitem[Geirhos et~al.(2018)Geirhos, Temme, Rauber, Sch{\"u}tt, Bethge, and
  Wichmann]{geirhos2018generalisation}
Robert Geirhos, Carlos~RM Temme, Jonas Rauber, Heiko~H Sch{\"u}tt, Matthias
  Bethge, and Felix~A Wichmann.
\newblock Generalisation in humans and deep neural networks.
\newblock In \emph{Advances in neural information processing systems}, pages
  7538--7550, 2018.

\bibitem[Goodfellow et~al.(2014)Goodfellow, Pouget-Abadie, Mirza, Xu,
  Warde-Farley, Ozair, Courville, and Bengio]{Goodfellow14GAN}
Ian Goodfellow, Jean Pouget-Abadie, Mehdi Mirza, Bing Xu, David Warde-Farley,
  Sherjil Ozair, Aaron Courville, and Yoshua Bengio.
\newblock Generative adversarial nets.
\newblock In Z.~Ghahramani, M.~Welling, C.~Cortes, N.~Lawrence, and K.~Q.
  Weinberger, editors, \emph{Advances in Neural Information Processing
  Systems}, volume~27. Curran Associates, Inc., 2014.
\newblock URL
  \url{https://proceedings.neurips.cc/paper/2014/file/5ca3e9b122f61f8f06494c97b1afccf3-Paper.pdf}.

\bibitem[Graves(2011)]{graves2011SVI}
Alex Graves.
\newblock Practical variational inference for neural networks.
\newblock In J.~Shawe-Taylor, R.~S. Zemel, P.~L. Bartlett, F.~Pereira, and
  K.~Q. Weinberger, editors, \emph{Advances in Neural Information Processing
  Systems 24}, pages 2348--2356. Curran Associates, Inc., 2011.

\bibitem[Gretton et~al.(2012)Gretton, Sejdinovic, Strathmann, Balakrishnan,
  Pontil, Fukumizu, and Sriperumbudur]{NIPS2012MKMMD}
Arthur Gretton, Dino Sejdinovic, Heiko Strathmann, Sivaraman Balakrishnan,
  Massimiliano Pontil, Kenji Fukumizu, and Bharath~K. Sriperumbudur.
\newblock Optimal kernel choice for large-scale two-sample tests.
\newblock In F.~Pereira, C.~J.~C. Burges, L.~Bottou, and K.~Q. Weinberger,
  editors, \emph{Advances in Neural Information Processing Systems}, volume~25.
  Curran Associates, Inc., 2012.
\newblock URL
  \url{https://proceedings.neurips.cc/paper/2012/file/dbe272bab69f8e13f14b405e038deb64-Paper.pdf}.

\bibitem[Guo et~al.(2017)Guo, Pleiss, Sun, and
  Weinberger]{guo_calibration_2017}
Chuan Guo, Geoff Pleiss, Yu~Sun, and Kilian~Q. Weinberger.
\newblock On {Calibration} of {Modern} {Neural} {Networks}.
\newblock In \emph{Proceedings of the 34th {International} {Conference} on
  {Machine} {Learning}, {ICML} 2017, {Sydney}, {NSW}, {Australia}, 6-11
  {August} 2017}, pages 1321--1330, 2017.

\bibitem[Gupta et~al.(2020)Gupta, Podkopaev, and Ramdas]{Gupta2020}
Chirag Gupta, Aleksandr Podkopaev, and Aaditya Ramdas.
\newblock Distribution-free binary classification: prediction sets, confidence
  intervals and calibration.
\newblock In H.~Larochelle, M.~Ranzato, R.~Hadsell, M.~F. Balcan, and H.~Lin,
  editors, \emph{Advances in Neural Information Processing Systems}, volume~33,
  pages 3711--3723. Curran Associates, Inc., 2020.
\newblock URL
  \url{https://proceedings.neurips.cc/paper/2020/file/26d88423fc6da243ffddf161ca712757-Paper.pdf}.

\bibitem[Hendrycks and Dietterich(2019)]{hendrycks2019robustness}
Dan Hendrycks and Thomas Dietterich.
\newblock Benchmarking neural network robustness to common corruptions and
  perturbations.
\newblock \emph{Proceedings of the International Conference on Learning
  Representations}, 2019.

\bibitem[Hendrycks and Gimpel(2017)]{hendrycks17baseline}
Dan Hendrycks and Kevin Gimpel.
\newblock A baseline for detecting misclassified and out-of-distribution
  examples in neural networks.
\newblock In \emph{Proceedings of International Conference on Learning
  Representations}, 2017.

\bibitem[Hendrycks et~al.(2019{\natexlab{a}})Hendrycks, Lee, and
  Mazeika]{hendrycks2019pretraining}
Dan Hendrycks, Kimin Lee, and Mantas Mazeika.
\newblock Using pre-training can improve model robustness and uncertainty.
\newblock \emph{Proceedings of the International Conference on Machine
  Learning}, 2019{\natexlab{a}}.

\bibitem[Hendrycks et~al.(2019{\natexlab{b}})Hendrycks, Mazeika, and
  Dietterich]{hendrycks2019oe}
Dan Hendrycks, Mantas Mazeika, and Thomas Dietterich.
\newblock Deep anomaly detection with outlier exposure.
\newblock \emph{Proceedings of the International Conference on Learning
  Representations}, 2019{\natexlab{b}}.

\bibitem[Hendrycks et~al.(2020)Hendrycks, Mu, Cubuk, Zoph, Gilmer, and
  Lakshminarayanan]{hendrycks2020augmix}
Dan Hendrycks, Norman Mu, Ekin~D. Cubuk, Barret Zoph, Justin Gilmer, and Balaji
  Lakshminarayanan.
\newblock {AugMix}: A simple data processing method to improve robustness and
  uncertainty.
\newblock \emph{Proceedings of the International Conference on Learning
  Representations (ICLR)}, 2020.

\bibitem[Jiang et~al.(2020)Jiang, Chen, Fu, and Long]{dalib}
Junguang Jiang, Baixu Chen, Bo~Fu, and Mingsheng Long.
\newblock Transfer-learning-library.
\newblock \url{https://github.com/thuml/Transfer-Learning-Library}, 2020.

\bibitem[Jin et~al.(2020)Jin, Wang, Long, and Wang]{Jin2020MCC}
Ying Jin, Ximei Wang, Mingsheng Long, and Jianmin Wang.
\newblock Minimum class confusion for versatile domain adaptation.
\newblock In Andrea Vedaldi, Horst Bischof, Thomas Brox, and Jan-Michael Frahm,
  editors, \emph{Computer Vision -- ECCV 2020}, pages 464--480, Cham, 2020.
  Springer International Publishing.

\bibitem[Krishnan and Tickoo(2020)]{AvUCloss}
Ranganath Krishnan and Omesh Tickoo.
\newblock Improving model calibration with accuracy versus uncertainty
  optimization.
\newblock In H.~Larochelle, M.~Ranzato, R.~Hadsell, M.~F. Balcan, and H.~Lin,
  editors, \emph{Advances in Neural Information Processing Systems}, volume~33,
  pages 18237--18248. Curran Associates, Inc., 2020.
\newblock URL
  \url{https://proceedings.neurips.cc/paper/2020/file/d3d9446802a44259755d38e6d163e820-Paper.pdf}.

\bibitem[Kull et~al.(2019)Kull, Perello~Nieto, K\"{a}ngsepp, Silva~Filho, Song,
  and Flach]{Kull2019beyond}
Meelis Kull, Miquel Perello~Nieto, Markus K\"{a}ngsepp, Telmo Silva~Filho, Hao
  Song, and Peter Flach.
\newblock Beyond temperature scaling: Obtaining well-calibrated multi-class
  probabilities with dirichlet calibration.
\newblock In H.~Wallach, H.~Larochelle, A.~Beygelzimer, F.~d\textquotesingle
  Alch\'{e}-Buc, E.~Fox, and R.~Garnett, editors, \emph{Advances in Neural
  Information Processing Systems}, volume~32. Curran Associates, Inc., 2019.

\bibitem[Lakshminarayanan et~al.(2017)Lakshminarayanan, Pritzel, and
  Blundell]{lakshminarayanan_simple_2017}
Balaji Lakshminarayanan, Alexander Pritzel, and Charles Blundell.
\newblock Simple and {Scalable} {Predictive} {Uncertainty} {Estimation} using
  {Deep} {Ensembles}.
\newblock In \emph{Advances in {Neural} {Information} {Processing} {Systems}
  30: {Annual} {Conference} on {Neural} {Information} {Processing} {Systems}
  2017, 4-9 {December} 2017, {Long} {Beach}, {CA}, {USA}}, pages 6405--6416,
  2017.

\bibitem[Lin et~al.(2014)Lin, Maire, Belongie, Hays, Perona, Ramanan,
  Doll{\'a}r, and Zitnick]{MicrosoftCOCO}
Tsung-Yi Lin, Michael Maire, Serge Belongie, James Hays, Pietro Perona, Deva
  Ramanan, Piotr Doll{\'a}r, and C.~Lawrence Zitnick.
\newblock Microsoft coco: Common objects in context.
\newblock In David Fleet, Tomas Pajdla, Bernt Schiele, and Tinne Tuytelaars,
  editors, \emph{Computer Vision -- ECCV 2014}, pages 740--755, Cham, 2014.
  Springer International Publishing.

\bibitem[Long et~al.(2015)Long, Cao, Wang, and Jordan]{Long15DAN}
Mingsheng Long, Yue Cao, Jianmin Wang, and Michael~I. Jordan.
\newblock Learning transferable features with deep adaptation networks.
\newblock In \emph{Proceedings of the 32nd International Conference on
  International Conference on Machine Learning - Volume 37}, ICML'15, pages
  97--105. JMLR.org, 2015.

\bibitem[Long et~al.(2017)Long, Zhu, Wang, and Jordan]{Long17JAN}
Mingsheng Long, Han Zhu, Jianmin Wang, and Michael~I. Jordan.
\newblock Deep transfer learning with joint adaptation networks.
\newblock In Doina Precup and Yee~Whye Teh, editors, \emph{Proceedings of the
  34th International Conference on Machine Learning}, volume~70 of
  \emph{Proceedings of Machine Learning Research}, pages 2208--2217. PMLR,
  06--11 Aug 2017.
\newblock URL \url{http://proceedings.mlr.press/v70/long17a.html}.

\bibitem[Long et~al.(2018)Long, CAO, Wang, and Jordan]{Long18CDAN}
Mingsheng Long, ZHANGJIE CAO, Jianmin Wang, and Michael~I Jordan.
\newblock Conditional adversarial domain adaptation.
\newblock In S.~Bengio, H.~Wallach, H.~Larochelle, K.~Grauman, N.~Cesa-Bianchi,
  and R.~Garnett, editors, \emph{Advances in Neural Information Processing
  Systems}, volume~31. Curran Associates, Inc., 2018.
\newblock URL
  \url{https://proceedings.neurips.cc/paper/2018/file/ab88b15733f543179858600245108dd8-Paper.pdf}.

\bibitem[Louizos and Welling(2016)]{louizos2016SVI}
Christos Louizos and Max Welling.
\newblock Structured and efficient variational deep learning with matrix
  gaussian posteriors.
\newblock In Maria~Florina Balcan and Kilian~Q. Weinberger, editors,
  \emph{Proceedings of The 33rd International Conference on Machine Learning},
  volume~48 of \emph{Proceedings of Machine Learning Research}, pages
  1708--1716, New York, New York, USA, 20--22 Jun 2016. PMLR.
\newblock URL \url{http://proceedings.mlr.press/v48/louizos16.html}.

\bibitem[Louizos and Welling(2017)]{louizos2017SVI}
Christos Louizos and Max Welling.
\newblock Multiplicative normalizing flows for variational {B}ayesian neural
  networks.
\newblock In Doina Precup and Yee~Whye Teh, editors, \emph{Proceedings of the
  34th International Conference on Machine Learning}, volume~70 of
  \emph{Proceedings of Machine Learning Research}, pages 2218--2227,
  International Convention Centre, Sydney, Australia, 06--11 Aug 2017. PMLR.
\newblock URL \url{http://proceedings.mlr.press/v70/louizos17a.html}.

\bibitem[Nado et~al.(2020)Nado, Padhy, Sculley, D'Amour, Lakshminarayanan, and
  Snoek]{nado2020evaluating}
Zachary Nado, Shreyas Padhy, D~Sculley, Alexander D'Amour, Balaji
  Lakshminarayanan, and Jasper Snoek.
\newblock Evaluating prediction-time batch normalization for robustness under
  covariate shift.
\newblock In \emph{ICML 2020 Workshop on Uncertainty and Robustness in Deep
  Learning}, 2020.

\bibitem[Nguyen et~al.(2015)Nguyen, Yosinski, and Clune]{nguyen_deep_2015}
Anh~Mai Nguyen, Jason Yosinski, and Jeff Clune.
\newblock Deep neural networks are easily fooled: {High} confidence predictions
  for unrecognizable images.
\newblock In \emph{{IEEE} {Conference} on {Computer} {Vision} and {Pattern}
  {Recognition}, {CVPR} 2015, {Boston}, {MA}, {USA}, {June} 7-12, 2015}, pages
  427--436, 2015.
\newblock \doi{10.1109/CVPR.2015.7298640}.
\newblock URL \url{https://doi.org/10.1109/CVPR.2015.7298640}.

\bibitem[Nixon et~al.(2019)Nixon, Dusenberry, Zhang, Jerfel, and
  Tran]{adaptiveECE}
Jeremy Nixon, Michael~W. Dusenberry, Linchuan Zhang, Ghassen Jerfel, and Dustin
  Tran.
\newblock Measuring calibration in deep learning.
\newblock In \emph{Proceedings of the IEEE/CVF Conference on Computer Vision
  and Pattern Recognition (CVPR) Workshops}, June 2019.

\bibitem[Ovadia et~al.(2019)Ovadia, Fertig, Ren, Nado, Sculley, Nowozin,
  Dillon, Lakshminarayanan, and Snoek]{SnoekPaper}
Yaniv Ovadia, Emily Fertig, Jie Ren, Zachary Nado, D.~Sculley, Sebastian
  Nowozin, Joshua Dillon, Balaji Lakshminarayanan, and Jasper Snoek.
\newblock Can you trust your model's uncertainty? {E}valuating predictive
  uncertainty under dataset shift.
\newblock In H.~Wallach, H.~Larochelle, A.~Beygelzimer, F.~d'Alch\'{e} Buc,
  E.~Fox, and R.~Garnett, editors, \emph{Advances in Neural Information
  Processing Systems 32}, pages 13991--14002. Curran Associates, Inc., 2019.

\bibitem[Park et~al.(2020)Park, Bastani, Weimer, and Lee]{ParkUDA}
Sangdon Park, Osbert Bastani, James Weimer, and Insup Lee.
\newblock Calibrated prediction with covariate shift via unsupervised domain
  adaptation.
\newblock In Silvia Chiappa and Roberto Calandra, editors, \emph{The 23rd
  International Conference on Artificial Intelligence and Statistics, {AISTATS}
  2020, 26-28 August 2020, Online [Palermo, Sicily, Italy]}, volume 108 of
  \emph{Proceedings of Machine Learning Research}, pages 3219--3229. {PMLR},
  2020.
\newblock URL \url{http://proceedings.mlr.press/v108/park20b.html}.

\bibitem[Peng et~al.(2017)Peng, Usman, Kaushik, Hoffman, Wang, and
  Saenko]{peng2017visda}
Xingchao Peng, Ben Usman, Neela Kaushik, Judy Hoffman, Dequan Wang, and Kate
  Saenko.
\newblock Visda: The visual domain adaptation challenge.
\newblock \emph{CoRR, abs/1710.06924}, 2017.

\bibitem[Rahimi et~al.(2020)Rahimi, Shaban, Cheng, Hartley, and
  Boots]{Rahimi2020intraorder}
Amir Rahimi, Amirreza Shaban, Ching-An Cheng, Richard Hartley, and Byron Boots.
\newblock Intra order-preserving functions for calibration of multi-class
  neural networks.
\newblock In H.~Larochelle, M.~Ranzato, R.~Hadsell, M.~F. Balcan, and H.~Lin,
  editors, \emph{Advances in Neural Information Processing Systems}, volume~33,
  pages 13456--13467. Curran Associates, Inc., 2020.

\bibitem[Recht et~al.(2019)Recht, Roelofs, Schmidt, and Shankar]{ImageNetV2}
Benjamin Recht, Rebecca Roelofs, Ludwig Schmidt, and Vaishaal Shankar.
\newblock Do {I}mage{N}et classifiers generalize to {I}mage{N}et?
\newblock In Kamalika Chaudhuri and Ruslan Salakhutdinov, editors,
  \emph{Proceedings of the 36th International Conference on Machine Learning},
  volume~97 of \emph{Proceedings of Machine Learning Research}, pages
  5389--5400. PMLR, 09--15 Jun 2019.
\newblock URL \url{https://proceedings.mlr.press/v97/recht19a.html}.

\bibitem[S\'emery(2021)]{imgclsmob}
Oleg S\'emery.
\newblock Face recognition using pytorch.
\newblock \url{https://github.com/osmr/imgclsmob}, 2021.

\bibitem[Shao et~al.(2020)Shao, Yang, and Ren]{shao2020calibrating}
Zhihui Shao, Jianyi Yang, and Shaolei Ren.
\newblock Calibrating deep neural network classifiers on out-of-distribution
  datasets, 2020.

\bibitem[Shimodaira(2000)]{Shimodaira2000}
Hidetoshi Shimodaira.
\newblock Improving predictive inference under covariate shift by weighting the
  log-likelihood function.
\newblock \emph{Journal of Statistical Planning and Inference}, 90\penalty0
  (2):\penalty0 227--244, 2000.
\newblock ISSN 0378-3758.
\newblock \doi{https://doi.org/10.1016/S0378-3758(00)00115-4}.
\newblock URL
  \url{https://www.sciencedirect.com/science/article/pii/S0378375800001154}.

\bibitem[Srivastava et~al.(2014)Srivastava, Hinton, Krizhevsky, Sutskever, and
  Salakhutdinov]{srivastava2014dropout}
Nitish Srivastava, Geoffrey Hinton, Alex Krizhevsky, Ilya Sutskever, and Ruslan
  Salakhutdinov.
\newblock Dropout: A simple way to prevent neural networks from overfitting.
\newblock \emph{Journal of Machine Learning Research}, 15:\penalty0 1929--1958,
  06 2014.

\bibitem[Vasiljevic et~al.(2016)Vasiljevic, Chakrabarti, and
  Shakhnarovich]{vasiljevic2016examining}
Igor Vasiljevic, Ayan Chakrabarti, and Gregory Shakhnarovich.
\newblock Examining the impact of blur on recognition by convolutional
  networks.
\newblock \emph{arXiv preprint arXiv:1611.05760}, 2016.

\bibitem[Venkateswara et~al.(2017)Venkateswara, Eusebio, Chakraborty, and
  Panchanathan]{OfficeHome}
Hemanth Venkateswara, Jose Eusebio, Shayok Chakraborty, and Sethuraman
  Panchanathan.
\newblock Deep hashing network for unsupervised domain adaptation.
\newblock In \emph{Proceedings of the IEEE Conference on Computer Vision and
  Pattern Recognition (CVPR)}, July 2017.

\bibitem[Wang et~al.(2019)Wang, Ge, Lipton, and Xing]{ImageNetSketch}
Haohan Wang, Songwei Ge, Zachary Lipton, and Eric~P Xing.
\newblock Learning robust global representations by penalizing local predictive
  power.
\newblock In \emph{Advances in Neural Information Processing Systems}, pages
  10506--10518, 2019.

\bibitem[Wang et~al.(2020)Wang, Long, Wang, and Jordan]{wang2020transferable}
Ximei Wang, Mingsheng Long, Jianmin Wang, and Michael~I. Jordan.
\newblock Transferable calibration with lower bias and variance in domain
  adaptation.
\newblock In \emph{Advances in neural information processing systems}, 2020.

\bibitem[Wen et~al.(2018)Wen, Vicol, Ba, Tran, and Grosse]{wen2018flipout}
Yeming Wen, Paul Vicol, Jimmy Ba, Dustin Tran, and Roger Grosse.
\newblock Flipout: Efficient pseudo-independent weight perturbations on
  mini-batches.
\newblock In \emph{International Conference on Learning Representations}, 2018.

\bibitem[Xu et~al.(2019)Xu, Li, Yang, and Lin]{Xu2019SAFN}
Ruijia Xu, Guanbin Li, Jihan Yang, and Liang Lin.
\newblock Larger norm more transferable: An adaptive feature norm approach for
  unsupervised domain adaptation.
\newblock In \emph{The IEEE International Conference on Computer Vision
  (ICCV)}, October 2019.

\bibitem[Zhang et~al.(2019)Zhang, Liu, Long, and Jordan]{Zhang2019MDD}
Yuchen Zhang, Tianle Liu, Mingsheng Long, and Michael Jordan.
\newblock Bridging theory and algorithm for domain adaptation.
\newblock In Kamalika Chaudhuri and Ruslan Salakhutdinov, editors,
  \emph{Proceedings of the 36th International Conference on Machine Learning},
  volume~97 of \emph{Proceedings of Machine Learning Research}, pages
  7404--7413. PMLR, 09--15 Jun 2019.
\newblock URL \url{http://proceedings.mlr.press/v97/zhang19i.html}.

\end{thebibliography}

\newpage
\appendix

{\Large{\textbf{Supplemental Material}}}

\section{CIFAR-10-C and ImageNet-C}

\begin{figure*}[ht]
\centering
\includegraphics[width=\linewidth]{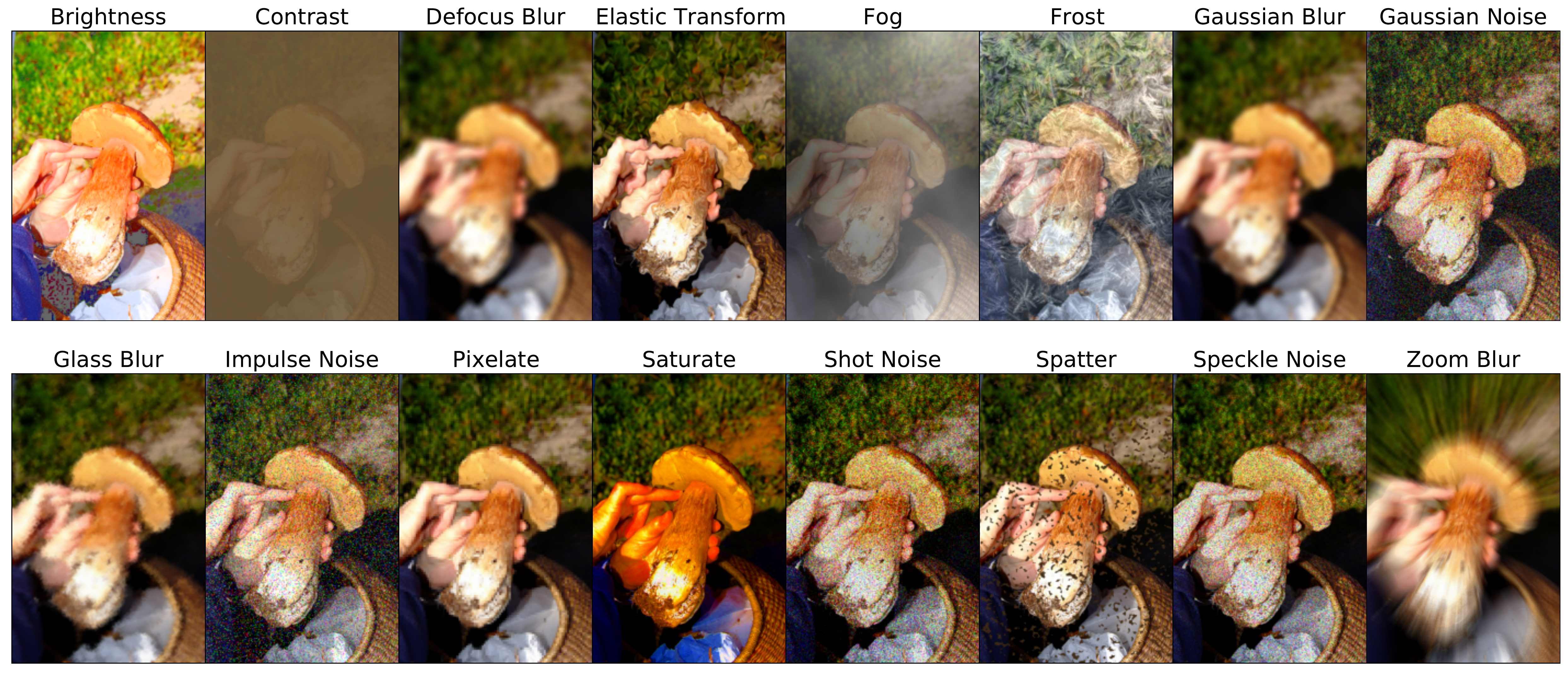}
\caption{Example of the different 16 corruptions used to form CIFAR-10-C and ImageNet-C \citep{hendrycks2019robustness}.}
\label{fig:corruptions}
\end{figure*}

\begin{figure*}[ht]
\centering
\includegraphics[width=\linewidth]{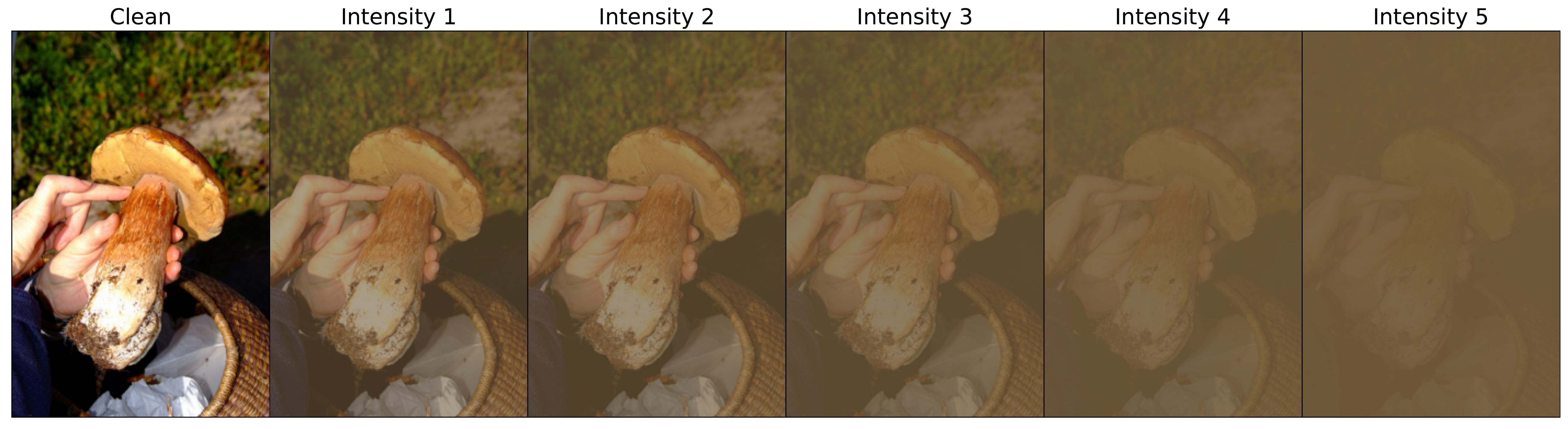}
\caption{Example of the contrast corruption at the different levels of intensity.}
\label{fig:corruptions_intensity}
\end{figure*}

CIFAR-10-C and ImageNet-C \citep{hendrycks2019robustness} are datasets obtained from the CIFAR-10 and ImageNet datasets, respectively, by applying common real-world corruptions at different levels of intensity. The corruptions considered in this work are: \emph{Brightness}, \emph{Contrast}, \emph{Defocus Blur}, \emph{Elastic Transform}, \emph{Fog}, \emph{Frost}, \emph{Gaussian Blur}, \emph{Gaussian Noise}, \emph{Glass Blur}, \emph{Impulse Noise}, \emph{Pixelate}, \emph{Saturate}, \emph{Shot Noise}, \emph{Spatter}, \emph{Speckle Noise}, \emph{Zoom Blur}. In \autoref{fig:corruptions} we show an example of all the different corruptions. In \autoref{fig:corruptions_intensity} we show the contrast corruption type in five different intensities.

\begin{figure*}[ht]
\centering
\includegraphics[width=0.5\linewidth]{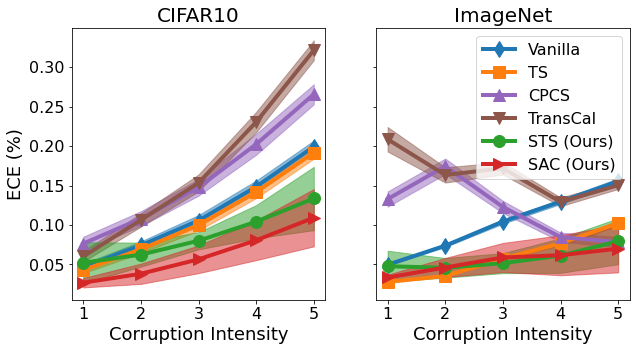}
\caption{Mean ECE (lower is better), averaged across different corruption types used in making the calibration sets. \autoref{fig:ECE_means} shows us the mean ECE using ``contrast'' as the calibration corruption. Here we show how those means change when different corruptions are used in the calibration set. For CIFAR-10, our proposed methods are robust to the choice of corruption used in the calibration set, while for ImageNet the choice of the corruption is more import, in particular for the Single Image method.}
\label{fig:ECEcorruption_analysis}
\end{figure*}

\section{Robustness to the choice of corruption}

We look how the choice of corruption to synthesize the surrogate calibration sets impacts our methods 
Ideally, the choice of corruption should be representative of the distribution of corruptions, so a mild corruption or a very strong corruption would give slightly worse results.  At the same time, here we demonstrate that choosing a different corruption should not significantly degrade the results.

In Figure \ref{fig:ECEcorruption_analysis} we perform a cross-validation study over the choice of corruption used to generate the calibration sets (always leaving it out of the corruptions used at test time). We plot the mean and variance of the ECE across different validation corruptions types. We find that for CIFAR-10, our AAC method is robust to the choice of corruption. For the ImageNet model, the robust remains although a careful choice may be necessary at higher levels of corruption.

\section{Variants of the SAC method}

\begin{figure*}[t]
\centering
\includegraphics[width=0.5\linewidth]{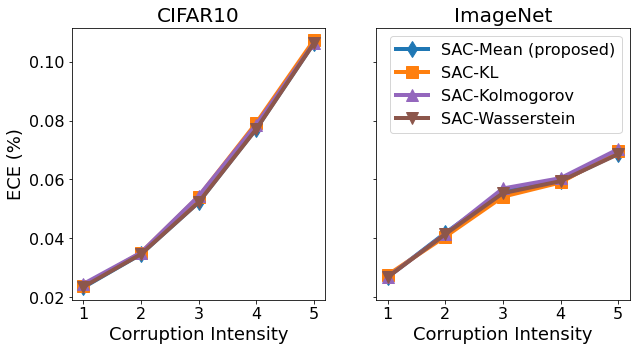}
\caption{Comparison of different distances to choose the surrogate calibration set: Mean Expected Calibration Error (ECE) (lower is better) of the benchmark implementation \citep{SnoekPaper}, versus our Multi-Image methods for ImageNet (top) and CIFAR-10 (bottom). Each box represents a different uncertainty method.}
\label{fig:multi_image_distance}
\end{figure*}

\begin{figure}[t]
\centering
\includegraphics[width=0.5\linewidth]{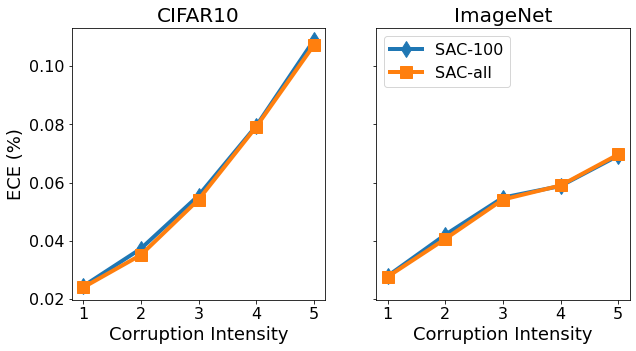}
\caption{Comparison of our Multi-Image method for ImageNet (top) and CIFAR-10 (bottom) using the full batch of test images and only 100 of them, which we refer to as peak.
Mean Expected Calibration Error (ECE) across different corruptions types, for fixed corruption intensity going from 0 to 5. Each box represents a different uncertainty method.}
\label{fig:multi_image_peak}
\end{figure}

As discussed in \autoref{sec:variants}, the SAC method is robust to the choice of distributional distance to choose the surrogate calibration set (see \autoref{fig:multi_image_distance}) and when using the mean only 100 samples are enough instead of the entire calibration set (see \autoref{fig:multi_image_peak}).

\section{Qualitative Results}

\begin{figure*}[h]
\centering
\includegraphics[width=\linewidth]{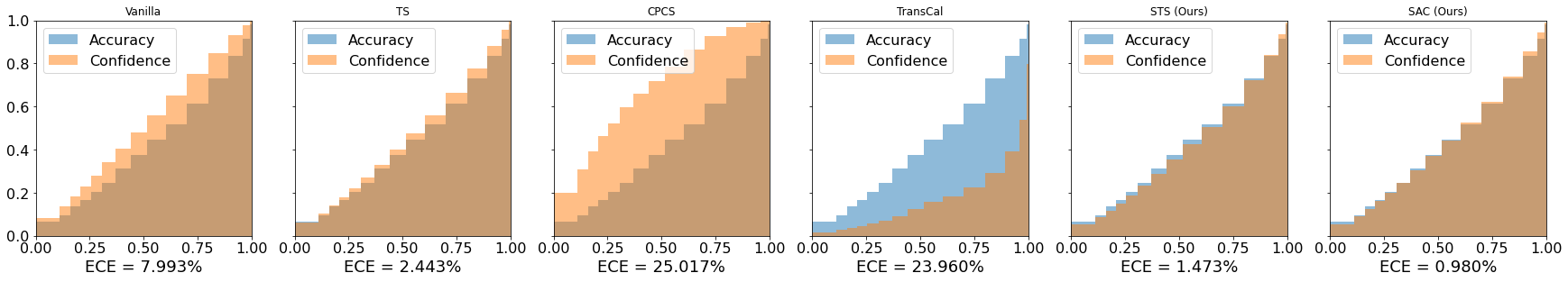}
\vspace{-.8em}
\caption{Calibration errors for the different methods on corrupted images on ImageNet using the elastic transform corruption with intensity 2.
The x-axis is the range of $\pmax$ values. 
We visualize as binned histograms the model accuracy (blue) (\autoref{eq:binned_accuracy}), and the confidence estimates (orange) (\autoref{eq:binned_confidence}) (brown is where they overlap).
The gap between the orange and blue curves represents the calibration error.
}
\label{fig:calibration_plots}
\end{figure*}

\begin{figure*}[t]
\centering
\includegraphics[width=\linewidth]{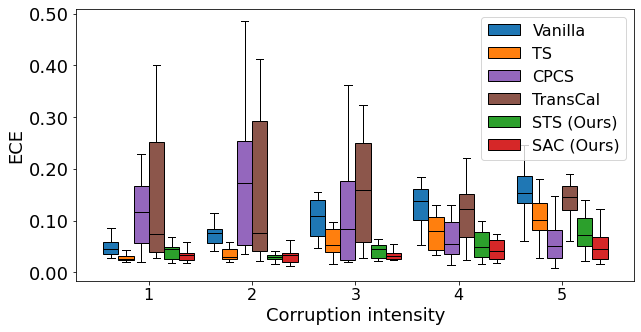}
\caption{Comparison of the ECE (lower is better) for across corruptions types for 5 corruption intensities using different calibration methods:  Vanilla, Temperature Scaling (TS), CPCS \citep{ParkUDA}, TransCal \citep{wang2020transferable}, our Surrogate Temperature Scaling (STS), and our Surrogate-Adaptive Calibration (SAC).
Our methods perform the best among all methods across all intensities, with the greatest improvement at higher intensities. For each method, we show the quartiles summarizing the results on each corruption intensity. The ECE using our approaches is consistently better.}
\label{fig:ECE_boxwhisker}
\end{figure*}

\autoref{fig:calibration_plots} shows the calibrated probabilities for a typical test set. \autoref{fig:ECE_boxwhisker} expands on \autoref{fig:ECE_means} by making use of a box-whisker plot to summarize the ECE across the different corruptions through its median and 1st and 3rd quantiles. 

\end{document}